\documentclass{article}

\usepackage[preprint]{neurips_2026}

\usepackage[utf8]{inputenc} 
\usepackage[T1]{fontenc}    
\usepackage{hyperref}       
\usepackage{url}            

\usepackage{booktabs}       
\usepackage{amsfonts}       
\usepackage{nicefrac}       
\usepackage{microtype}      
\usepackage{xcolor}         
\usepackage{graphicx}
\usepackage{subcaption}

\usepackage{xcolor}
\usepackage{hyperref}

\definecolor{papertan}{HTML}{6B4E8C}
\definecolor{paperpurple}{HTML}{B0713A}

\hypersetup{
    colorlinks=true, 
    linkcolor=paperpurple, 
    citecolor=paperpurple, 
    urlcolor=papertan, 
}

\title{Modelpedia: A Catalog of Model\\ Findings for the Meta-Science of AI}

\author{%
Franciszek Bernat$^{1,2}$\thanks{) Equal contribution} \quad
Dawid Płudowski$^{1,2,*}$ \quad
\textbf{Michał Jan Włodarczyk}$^{1,2}$ \\
\textbf{Luca Longo}$^{3}$ \quad
\textbf{Jianlong Zhou}$^{4}$ \quad
\textbf{Andreas Holzinger}$^{5}$ \quad
\textbf{Riccardo Guidotti}$^{6,7}$ \\
\textbf{Wojciech Samek}$^{8,9,10}$  \quad
\textbf{Przemysław Biecek}$^{1,2,11}$ \\[2pt]
$^{1}$Centre for Credible AI \quad
$^{2}$Warsaw University of Technology \quad
$^{3}$University College Cork \\
$^{4}$University of Technology Sydney \quad
$^{5}$Human-Centered AI Lab \quad
$^{6}$University of Pisa \\
$^{7}$ISTI-CNR \quad
$^{8}$Technical University of Berlin \quad
$^{9}$Fraunhofer Heinrich Hertz Institute \\
$^{10}$Berlin Institute for the Foundations of Learning and Data (BIFOLD) \\
$^{11}$University of Warsaw \\
}

\begin{document}

\maketitle

\begin{abstract}
  Scientific knowledge about AI models is produced faster than the community can organize it. Every few months a new foundation model reshapes the field and hundreds of papers, blogs, and technical reports document how each behaves or fails. Yet, these findings remain scattered and effectively unretrievable.
To address this gap we present Modelpedia${}^1$, an automated, LLM-assisted framework that extracts findings about models from published papers, links it to the model, dataset, method, and concept it concerns, and aggregates the result into a searchable public catalog. Applying the prototype to accepted ICLR 2024 and 2025 papers, we extract over a thousand findings and, treating the catalog itself as an object of study, run a meta-analysis of how the community investigates models.
Now, we invite the community to explore, contribute to, and build on the open catalog, and to help establish model findings as a shared foundation for the meta-science of AI.


\footnotetext[1]{the service is available under the following link: \href{https://credibleai.github.io/modelpedia}{https://credibleai.github.io/modelpedia}, along with the source code: \href{https://github.com/CredibleAI/modelpedia}{https://github.com/CredibleAI/modelpedia}.}


\end{abstract}

\section{Introduction}

\setcounter{footnote}{1}

Foundation models achieve a remarkable performance in many vision~\citep{oquab2023dinov2}, language~\citep{team2025gemma}, and vision-language~\citep{tschannen2025siglip} tasks. 
As deployment has scaled, so has the literature on these models: paper after paper documents how a given model behaves, why it fails, and which shortcuts, biases, or hidden mechanisms drive its predictions.
This shift motivates what has recently been framed as \emph{Model Science}: the argument that the community should move beyond benchmarking new models and instead study existing ones systematically, treating a~trained model as an object of empirical investigation in its own right~\citep{biecek2026case}.

However, this knowledge is effectively locked away.
It is scattered across thousands of individual papers and internal reports; it is phrased inconsistently. 
There is no place where an observation about ``model~$X$'' accumulates, connects to related observations about other models, or can be retrieved by anyone asking a simple question such as \emph{``what do we know about how this model behaves?''}.
When a new model appears, the community moves on, and hard-earned findings about the previous one are rarely consolidated or reused. 
The result is an illusion of progress -- we produce more knowledge about models than ever before, but we cannot easily find it, compare it, or build on it.
This creates a~serious challenge for the scientific community to understand the specific model, as existing forms of meta-analysis do not address the need for summaries about models' findings.

As an example, surveys are still the most established form of meta-analysis in machine learning.
However, they focus on the broader research area rather than a single model.
While some notable examples, such as~\citep{mao2024gpteval, li2026clip}, exist, they are focused on the application of models, rather than investigating their behavior.
Benchmarks like LiveCodeBench or LiveBench~\citep{jain2025livecodebench, white2025livebench} are constantly updated and track how models perform over time, but do not go deeper into the models' internals.


Compared to those meta-analysis approaches, 
we envision a different regime, in which such findings about models are collected
in a \emph{unified} and \textit{structured} form. Just as model cards
\citep{mitchell2019modelcards}, datasheets \citep{gebru2021datasheets}, and data cards \citep{pushkarna2022datacards} introduced shared templates that made documentation of models and datasets comparable and reusable, we argue that \emph{findings} about models deserve their own unified representation.

\begin{figure}[t]
    \centering
    \includegraphics[trim=9cm 3.5cm 9cm 3.5cm, clip,width=\linewidth]{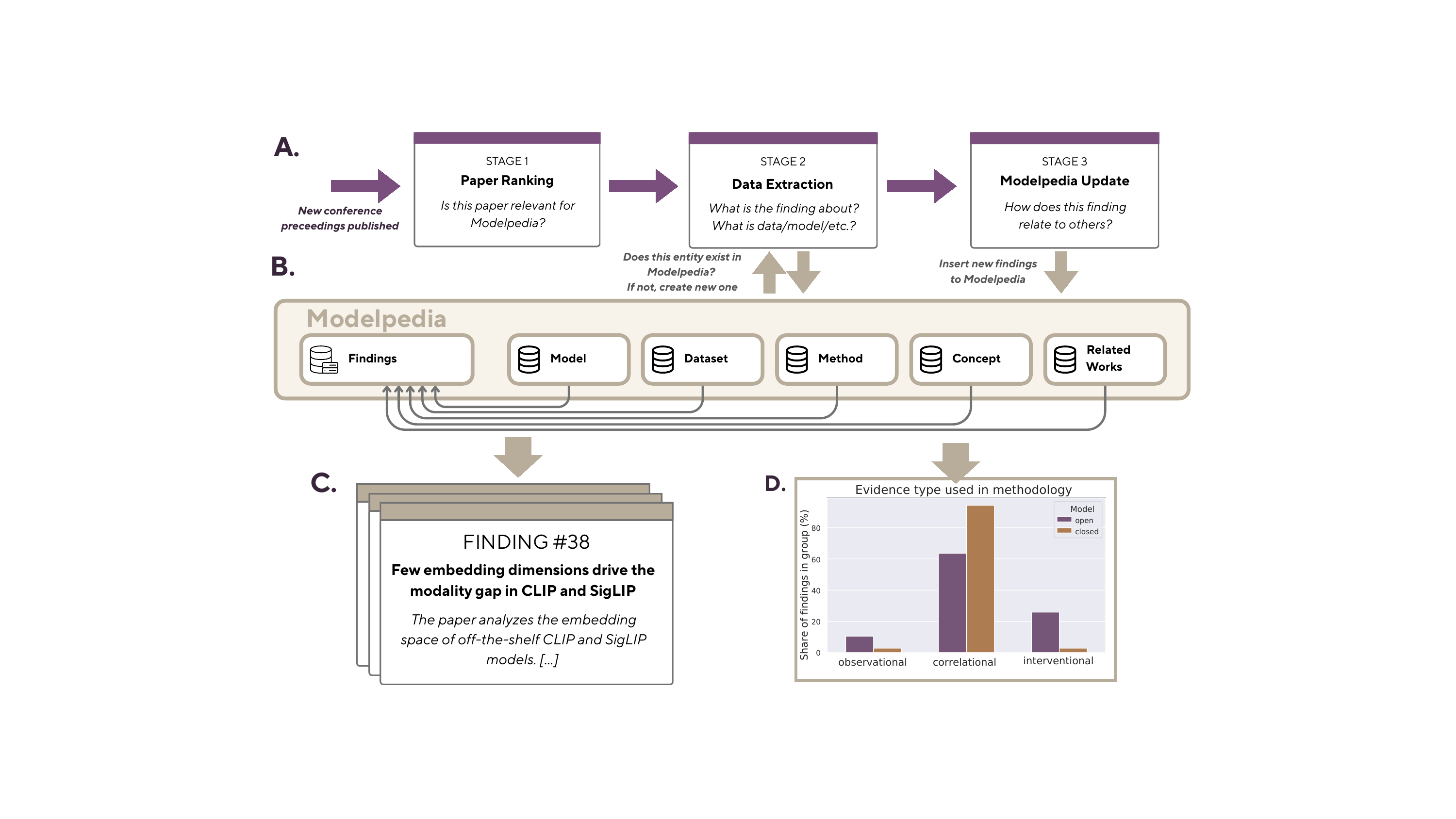}
    \caption{
    The overview of the Modelpedia framework.
    First, we capture articles that may contain findings about the model from conferences' proceedings (A). 
    Then, we put them into the structured database to ensure that each finding is supported by additional information, such as which model it describes and what data it was evaluated (B).
    The database is used to create summaries about each finding served via the web interface, allowing to complex analysis of models behavior (C).
    Finally, we perform a meta-analysis of extracted findings to better understand what is important for the community (D). 
    }
    \label{fig:figure-one}
\end{figure}

This paper presents {Modelpedia}, a first prototype implementation of this vision. Modelpedia is an~automated, LLM-assisted framework that extracts findings about foundation models from published papers and aggregates them into a single knowledge catalog. 
The overview of our methodology is presented in Figure~\ref{fig:figure-one}.
Our prototype already links findings to the models extracted from papers accepted to ICLR 2024 and 2025.
 
\textbf{Contributions.} (1)
We introduce the conceptual design of Modelpedia as a knowledge catalog of model findings. 
We formulate the idea of collecting third-party findings about models in a unified, structured database and position it as a missing piece of the Model Science agenda and of meta-science for artificial intelligence more broadly.
(2) A prototype implementation. 
We design and release an automated, LLM-assisted pipeline that extracts, verifies, links, and aggregates findings from published papers into a searchable knowledge catalog. Additionally, we present the web service that allows researchers to access the catalog easily.
(3) A meta-analysis of the catalog, characterizing which models and benchmarks concentrate the community’s attention.

\section{Modelpedia Architecture}




This section describes our prototype implementation of Modelpedia.
First, we present the database structure that supports analyzing findings about models.
Then, we discuss the process of extracting the findings.

\textbf{The database structure.}
We argue that each finding about the model should be supported with at least five references: (1) the \emph{model variant} for which the finding is provided, (2) the \emph{dataset} on which it was evaluated, (3) the \emph{method} that was used to make the finding, (4) the \emph{concept} that it represents, and (5) the \emph{source} being the original work.
Here, we use the term \emph{concept} to denote the general category of the finding, e.g., `failure mode' or `shortcut'.
Additionally, links to (6) \emph{related works} may be included, linking the finding to connected papers in the field.

The proposed structure implies a database with one main table containing findings and five related tables.
We present the detailed view of each of these tables in Appendix~\ref{app:data-structure}.
and provide an example of the entry in Appendix~\ref{app:master-table-example}.
The high-level structure of the database is also presented in Figure~\ref{fig:figure-one}\textcolor{paperpurple}{B}.

\textbf{The findings extraction.}
We base our prototype on the OpenReview system, which provides an API for extracting articles and hosts the most important machine learning conferences. 
For a prototype, we use Qwen3.8-27B and restricted the extraction to the ICLR conference.
The high-level description of the extraction is presented in Figure~\ref{fig:figure-one}\textcolor{paperpurple}{A}.

We note that only a fraction of articles in each conference provide findings about models, while most of them are focused on proposing new methodologies. 
To keep the computation manageable, we rank articles with a lightweight scoring system that does not use an LLM and relies solely on abstracts and reviews to evaluate article relevance. 
We use only approximately the top 15\% relevant articles for further analysis.
The details of scoring algorithms are presented in Appendix~\ref{app:scoring-system}.
As presented in Figure~\ref{fig:meta-analysis-1}, the top 15\% selection provides a reasonable trade-off between computational cost and relevance, with approximately 45\% of the selected articles subsequently found to lack any finding, as in Figure~\ref{fig:meta-analysis-3}\textcolor{paperpurple}{B}.
We highlight that the proposed scoring serves only as a guideline for the order of analysis of articles. 
Eventually, we plan to perform a complete analysis of all articles for the given conference.

Next, each extracted article is converted into a text file and used in a prompt to the LLM to find potential findings about models. 
The prompt is designed to enforce the structure of `entity': `quote from the article' to ensure it is not hallucinated.
The only exception to the rule being \emph{concept}, which we acknowledge might not be directly indicated by authors (e.g., authors may not state that their finding is about `failure mode' of the model).

To ensure that related tables do not contain duplicates (e.g., `GPT-4o' and `ChatGPT-4o', which are clearly the same model), we first use the LLM to match the extracted entity to already existing data. 
When matching is not possible---the entity does not yet exist in the database---the LLM proposes a new record in the database to update it.
We provide the extended information about the LLMs we used, along with the prompts, in Appendix~\ref{app:llms}.
Ultimately, all extracted records are used to update the database, ensuring correct linking between the main table and related tables.

In the end, we captured 1026 findings from two conferences. To ensure their correctness, we validate them with Claude Opus 5 with extra thinking. The validator LLM confirms all findings, with a small discrepancy about the `key metric' field on approximately 20\% of records.

\section{The case for Modelpedia}

\begin{figure}[t]
    \centering
    \begin{subfigure}[b]{0.48\textwidth}
        \centering
        \includegraphics[width=\textwidth]{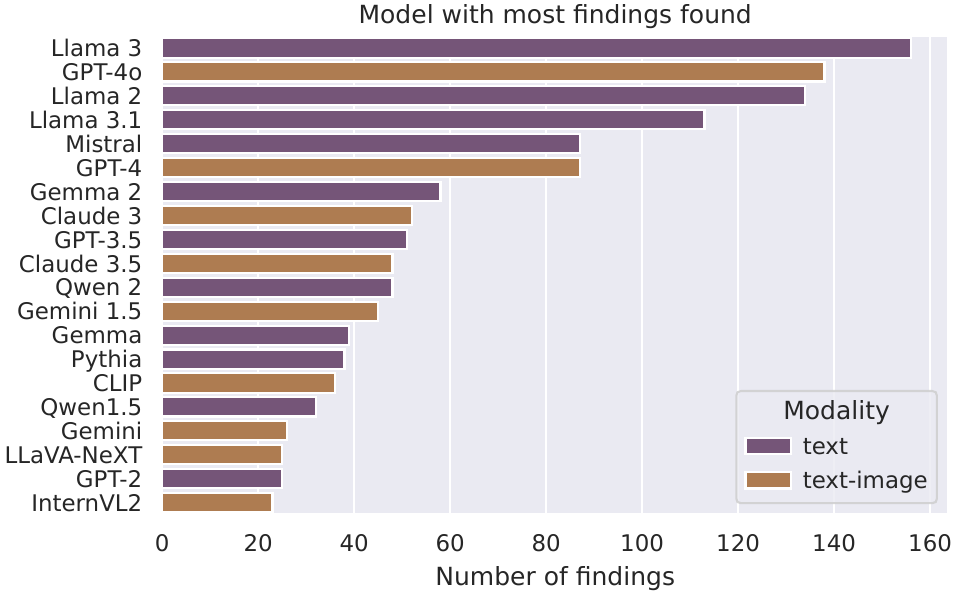}
    \end{subfigure}
    \hfill
    \begin{subfigure}[b]{0.48\textwidth}
        \centering
        \includegraphics[width=\textwidth]{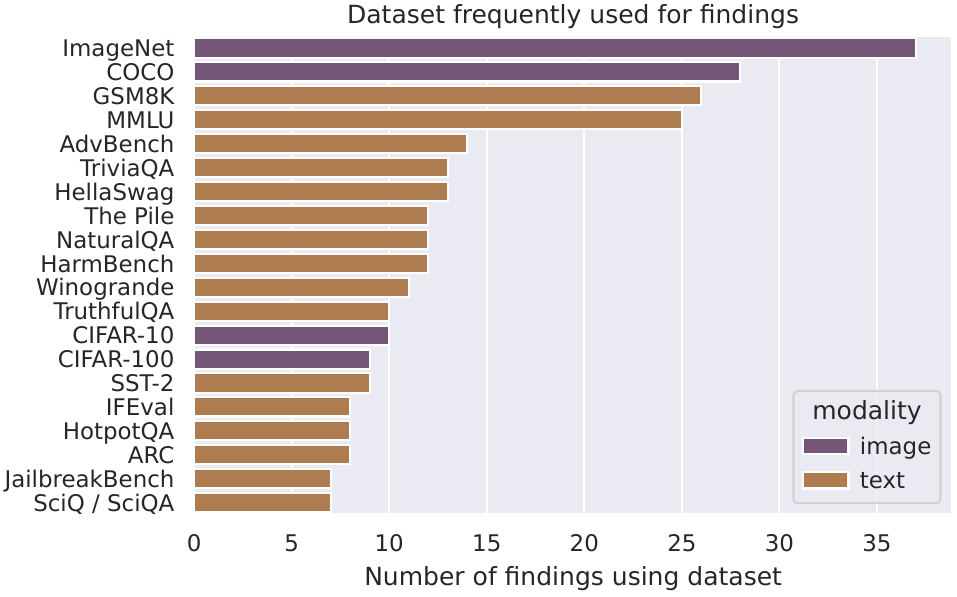}
    \end{subfigure}
    \caption{Meta-analysis of models (left) and datasets (right) used for making findings in articles in ICLR 2025. Language and vision-language models are the most analyzed. Interestingly, vision-based evaluation is mostly performed on the same datasets -- ImageNet and COCO, while language-based benchmarks are much more diverse.}
    \label{fig:meta-analysis-main}
\end{figure}


In this section, we perform a short meta-analysis of extracted findings to better understand the direction of the research community. 
Presented results are predominantly based on articles from ICLR 2025.
We plan to extend the analysis to other conferences in the future.

We found that most of the articles report findings based on correlational evidence, while observational or interventional evidence is used only by a minority of researchers (Figure~\ref{fig:figure-one}\textcolor{paperpurple}{D}).
This is a concerning result; correlation alone tells us little about why a model behaves the way it does.
The reliance on correlational evidence is common across both closed- and open-source models.
However, most findings based on observational or interventional evidence rely on open-source models. 
This is unsurprising, as interventions often require access to model weights. 
At the same time, it highlights the need for greater emphasis on analyzing open-source models, which allow for more direct investigation of model behavior.

Further, we evaluate which models are analyzed the most by the community. As presented in Figure~\ref{fig:meta-analysis-main} (left), the language and vision-language modalities are by far the most analyzed by researchers, with a strong preference for LLMs from Llama, GPT, and Mistral families.
Many findings also concern closed-source models like GPT, which, by construction, restricts researchers to mostly correlational evidence.

The analysis of the datasets used for creating findings highlights a strong preference of vision models towards well-established benchmarks, such as ImageNet~\citep{deng2009imagenet} and COCO~\citep{lin2014microsoft}. 
Language models, by contrast, are evaluated on a far more varied set of benchmarks.
As pointed out by~\citet{wang2024sober}, some bias evaluation of models like CLIP~\citep{radford2021learning} are heavily influenced by the inherent biases in the ImageNet dataset, leaving other critical aspects of the bias unexplored.
We provide an extended meta-analysis using Modelpedia in Appendix~\ref{app:extended-analysis}, where, among other points, we take into consideration how many findings a typical article produces, and how many datasets are used to evaluate a finding.





\section{Discussion}


Modelpedia rests on a simple claim: findings about models are first-class, creditable units of knowledge. Right now, an observation about a model only gets recognized if it becomes a full paper. The same observation tucked into an appendix is effectively lost, and that skews what the field thinks it knows, because the visible record reflects what was publishable, not what was learned. If findings become citable, versioned, and attributable, credit can go to whoever first documented it. Modelpedia is a catalog of such findings.

A catalog like this gets more useful the bigger it grows. As findings from more conferences pile up, Modelpedia stops being a passive record and becomes a tool for meta-analysis. It can show what the field has learned as well as what it hasn't studied. It can detect blind spots we can already see, like the field’s reliance on correlational over interventional evidence. In time, it could even help steer research toward the experiments that would tell us the most.

We see current version of Modelpedia as a starting point of a shared infrastructure for meta-science of AI models, and we invite the community to extract findings from more venues, add and curate entries, and build on the catalog, so that what we learn about models adds up over time instead of being rewritten and forgotten.




At last, we note that Modelpedia presents findings `as-is' in the sense that we do not evaluate whether the methodologies in articles are correct, and restrict ourselves to referring to their claims. 
Verifying whether published findings actually hold is a problem the community still has to solve; Modelpedia does not solve it, but it makes it tractable.

\acksection
Work on this project is financially supported by the Foundation for Polish Science (FNP) grant
‘Centre for Credible AI’ No. FENG.02.01-IP.05-0058/24.




\bibliographystyle{plainnat}
\bibliography{references}

\clearpage
\appendix

\section{Data structure}
\label{app:data-structure}

In this section, we provide a detailed view of the data structure we proposed for Modelpedia. The description stated below is presented in Figure~\ref{fig:figure-one}.


We begin by defining the shared information nodes that constitute each finding. These entities include: (1) \emph{Models}, the foundational node encompassing considered variants and analyzed behaviors, as findings are inherently tied to a model; (2) \emph{Datasets} utilized to derive the finding; (3) \emph{Methods} applied to evaluate model behavior and test the authors' claims; (4) \emph{Source}, which references the original paper for additional context; and (5) \emph{Concept}, an optional, generalized category classifying the model's behavior.

We believe that those five aspects are the core foundation of every finding. To streamline retrieval, nodes (1)-(4) are equipped with an anchor and an artifact. 
An anchor is a link to a paper where the entity was presented, if there is one. 
An artifact is a link to a site, where a model or a dataset can be downloaded. 
We highlight that it is not always possible to provide a link since not all datasets and models are easily accessible.

A final link records (6) \emph{related work} -- the prior methods, models or papers a given study positions itself against. It serves one of the three roles: 'builds-on' -- if the finding is an extension of what was presented in a related paper, 'compared-to' -- if the finding is being compared or contrasted with a claim from a related study or 'context' -- if the related work is essential to comprehend the full scope of the finding. \emph{Related work} entries, unlike those of the other fields, are rarely shared: an outside work cited here scarcely ever reappears elsewhere, and so, it does not have a node of its own. It is therefore kept inline, as a title and a link, which is sufficient to preserve the trail to the outside literature rather than being dropped.

Next, we define a set of fields designed to facilitate rapid comprehension, allowing users to evaluate a finding without referring to the original paper. The (i) \emph{Title} provides a concise, one-sentence description of the claim, serving as a primary entry point. Since a brief title may omit crucial nuances, it is supplemented by a (ii) \emph{Short summary} that overviews the finding and its underlying methodology. The (iii) \emph{Caveat} field details specific characteristics, including noteworthy author observations. To quantify the empirical reasoning behind a claim, the (iv) \emph{Key metric} documents the central measurement used to validate it. Additionally, an (v) \emph{Extraction flag} indicates whether the finding was extracted automatically or manually.

Finally, the (vi) \emph{Evidence type} is split into three main categories: 'Observational' -- a pattern is described as seen, without any measurements against a reference; reference (e.g., an attribution map that highlights the same region on every input). 'Interventional' -- something in the model that was deliberately changed led to significantly different output (e.g., steering a direction in the residual stream or truncating a weight matrix and recording the effect). 'Correlational' -- a quantity in a model is measured and compared across models, layers, sizes, or conditions, but nothing inside the model was altered (e.g., a metric that changes in different layers).

The ontology described above was formulated after a close reading of several manually selected papers containing detailed descriptions of exemplary findings and was informed by our own experience in explainable AI.
Those seed-round papers produced the first records in the database, which carry the 'manual-extraction' flag.

\clearpage

\section{Example of Finding}
\label{app:master-table-example}

Here, we provide an example of the entry for the finding in our database.\footnote{The ordering of individual components of an entry has been reshuffled here compared to Appendix~\ref{app:data-structure}, and resembles the read-through order a Modelpedia user will encounter.}
The finding describes the bias of the model from the CLIP family, which exhibits a strong bias towards textual information on the image~\citep{baniecki2026explaining}.
We also note that the presented example can be found on the Modelpedia web service.\footnote{\url{https://credibleai.github.io/modelpedia/findings/FX-003/index.html}}

As presented on the Modelpedia site, every finding page begins with a title, the authors of the paper that produced this finding, and a source where the paper can be accessed. After clicking on the source, you are transferred to a page with the date of the paper's submission, the venue to which the paper was submitted, the anchor, the artifact, and occasionally a note if necessary. In case of~\citet{baniecki2026explaining}, it is accordingly: date -- 2025-11-18, venue -- NeurIPS 2025, anchor -- link to arXiv\footnote{\url{https://arxiv.org/abs/2508.05430v2}} with paper, and artifact -- link to GitHub repository\footnote{\url{https://github.com/hbaniecki/fixlip}} relevant to the project.

Next, three things are relevant to describe what the finding is about. Those fields are without any handle to an outside work: (1) \textbf{Short summary}: \emph{In one qualitative CLIP explanation, FIXLIP assigns the strongest interaction to the caption token doll and an image patch containing the printed word dollar. The authors describe this as a case where one could say the model is right for the wrong reasons.} (2) \textbf{Evidence type}: \emph{Observational}. (3) \textbf{Caveat}: \emph{This is one qualitative interaction map. The paper reports neither its frequency nor a targeted intervention removing the printed word, so it supports a local explanation of this prediction, not a general causal claim that printed text drives CLIP matching.}

After that, the following fields are present: (4) \textbf{Model}: \emph{
CLIP}. (5) \textbf{Concept}: \emph{Shortcut, Feature interaction}. (6) \textbf{Method}: \emph{Weighted Banzhaf interaction index}. (7) \textbf{Related work}: \emph{
Eyes Wide Shut? Exploring the visual shortcomings of multimodal LLMs}. The argument for including those fields is provided in Appendix~\ref{app:data-structure}. By design, every node holds its own page on Modelpedia. On the page, the user is supplied with a summary about the referenced node and an optional note, if needed.

 The \textbf{Model} page also has a \textbf{variants} section listing models that belong to the same family. Also listed is the information about the developer, the date when the model was released, the modality, and the task the model was created to solve.

The final two records are: (8) \textbf{Related findings}: \emph{FX-001} -- present if more than one finding was extracted from the same source -- and (9) \textbf{Extraction}: \emph{automatic-extraction}.

Note that this entry lacks the \textbf{Dataset} and \textbf{Key metric} fields. Although the source paper clearly references a dataset, the extraction for this finding focuses on a single case. Hence, the dataset is omitted here, even though the dataset appears in a separate finding, extracted from the same paper. Since the nature of the finding is predominantly qualitative, there is no metric. A detailed explanation of how the findings are extracted is in Appendix~\ref{app:llms}.
\clearpage
Here, we provide an example of a final state in which the findings are presented in the database after the whole extraction, based on the finding described above.
\begin{verbatim}
id: FX-003
extracted_by: automatic-extraction
title: FIXLIP's strongest interaction in one CLIP example links doll to an
image patch reading dollar
description: >
  In one qualitative CLIP explanation, FIXLIP assigns the strongest
  interaction to the caption
  token doll and an image patch containing the printed word dollar.
  The authors describe this as
  a case where one could say the model is right for the wrong reasons.
evidence_type: observational
concepts:
  - ref: concept:shortcut
  - ref: concept:feature-interaction
key_metric: null
caveat: >
  This is one qualitative interaction map. The paper reports neither its
  frequency nor a targeted
  intervention removing the printed word, so it supports a local
  explanation of this prediction,
  not a general causal claim that printed text drives
  CLIP matching.
models:
  - ref: model:clip
    variant: not-specified-in-source
sources:
  - ref: source:fixlip
datasets: []
methods:
  - ref: method:weighted-banzhaf
    role: primary
related_work:
  - name: "Eyes Wide Shut? Exploring the visual shortcomings
    of multimodal LLMs"
    anchor: https://arxiv.org/abs/2401.06209
    role: context
related_findings:
  - FX-001
\end{verbatim}

\clearpage
\section{Scoring algorithm}
\label{app:scoring-system}

To populate the database, we downloaded metadata from ICLR 2024 and ICLR 2025 using the OpenReview API. 
Given the volume of submissions and the limits of the API, we restricted ourselves to accepted papers.
Our scoring system looks for keywords across the metadata: the abstracts and the reviews. After the algorithm completed, we downloaded the PDFs that scored $\geq 8.0$ points. We extracted a total of 972 papers: 451 from ICLR 2025 and 521 from ICLR 2024.

Those keywords are classified into seven groups with attached weights $w$.
Each group contribution is defined as $w \cdot \min(m, c)$, where $m$ is the number of distinct keywords matched. To address the issue of one specific group marginalizing other ones, we apply $c$ as a per-group ceiling. In addition, a keyword found in the reviews counts only when at least half of the reviewers of that paper used it independently. Authors and reviewers describe the same paper from different perspectives, so the vocabularies must slightly differ for abstracts and reviews, to distinguish phrases such as \emph{``We propose``} from \emph{``The paper proposes``}. Reviews are longer and come from several people, so they need higher ceilings, and the agreement gate is what keeps the extra matches signal rather than noise. The groups and their scoring are presented in Table~\ref{tab:scoring}. Exact vocabularies, weights, and thresholds used in the script are in the repository.\footnote{\url{https://github.com/CredibleAI/modelpedia/blob/main/modelpedia/ingest/screen.py}}

\begin{table}[h]
\centering
\caption{The seven keyword vocabularies, with the weight $w$ and the per-group
ceiling $c$ applied on each part of the score. A dash marks a vocabulary that
part does not use. Terms are matched either as prefixes (\emph{hallucinat} covers \emph{hallucinate} and
\emph{hallucination}) or as whole words. There are 30-60 keywords for each group. Eight behavior terms, including \emph{limitation}, \emph{robustness}
and \emph{bias}, are removed from the review vocabulary, since a reviewer applies
them to almost any paper.
\vspace{0.1cm}
}

\label{tab:scoring}
\small
\setlength{\tabcolsep}{4pt}
\begin{tabular}{llcccc}
\toprule
& & \multicolumn{2}{c}{Abstract} & \multicolumn{2}{c}{Reviews} \\
\cmidrule(lr){3-4} \cmidrule(lr){5-6}
Vocabulary & Example terms & $w$ & $c$ & $w$ & $c$ \\
\midrule
XAI & \emph{shap}, \emph{logit lens}, \emph{activation patching} & $2.0$ & $2$ & $2.0$ & $3$ \\
Model & \emph{clip}, \emph{llama-3}, \emph{stable diffusion} & $2.0$ & $2$ & $1.5$ & $2$ \\
Behavior & \emph{shortcut}, \emph{spurious}, \emph{hallucinat} & $1.0$ & $2$ & $1.0$ & $2$ \\
\midrule
Finding & \emph{we find}, \emph{our analysis}, \emph{surprisingly} & $1.0$ & $2$ & -- & -- \\
Method & \emph{we propose}, \emph{our framework}, \emph{outperforms} & $-0.5$ & $2$ & -- & -- \\
\midrule
Analysis & \emph{reveals that}, \emph{dissect}, \emph{pitfall} & -- & -- & $1.0$ & $4$ \\
Proposes & \emph{proposed method}, \emph{novel framework} & -- & -- & $-2.0$ & $3$ \\
\bottomrule
\end{tabular}
\end{table}

The abstracts and the reviews are scored separately, producing two independent rankings, which we merge by summing the two scores. The ranking of the five papers that scored the highest is visible in Table~\ref{tab:ranking-iclr2024}.

\begin{table}[h]
\centering
\caption{
The five highest-scoring accepted papers from ICLR 2024, with the two halves of the score shown separately.
\vspace{0.1cm}
}
\label{tab:ranking-iclr2024}
\small
\begin{tabular}{clrrr}
\toprule
& Paper & Abstract & Reviews & Total \\
\midrule
1 & Vision Transformers Need Registers                        & 9.5  & 14.0 & 23.5 \\
2 & Language Models Represent Space and Time                  & 9.0  & 11.5 & 20.5 \\
3 & Successor Heads: Recurring, Interpretable Attention Heads & 10.0 & 9.0  & 19.0 \\
4 & In-Context Learning Dynamics with Random Binary Sequences & 9.0  & 9.5  & 18.5 \\
5 & The Unlocking Spell on Base LLMs                          & 8.5  & 10.0 & 18.5 \\
\bottomrule
\end{tabular}
\end{table}

\clearpage
\section{LLM prompts}
\label{app:llms}

In this section, we describe how LLMs were used in the extraction process. We use {Qwen/Qwen3.8-27B-FP8} with thinking set to \emph{medium}. Four prompt templates are used across three steps. Every template shares one property: the model never invents an identifier that the entities are connected with. It either chooses from a closed list supplied in the prompt or states nothing and says why. 

The steps are separated by deterministic, non-LLM stages. Between Step I and Step II, every citation the model wrote is checked against the source by content-word overlap, and candidate names are matched against the registry by string-similarity. Between Step II and Step III, the accepted entities are written into the registry file, and anchors are derived from the citations.

\subsection{Step I: Extraction}
\label{step_I}
After the conversion of a paper into a text file, one prompt per paper (all prompt templates are available in the subfolder of the project repository\footnote{\url{https://github.com/CredibleAI/modelpedia/blob/main/modelpedia/ingest}}) is generated, containing the whole content from the source material, converted into text. The
prompt defines a finding as a \emph{third-party, post-hoc claim about a specific, pre-existing model}, which is the test that excludes claims about models the authors trained themselves. It
enforces the record structure, and orders the model to attach a citation from the source for every entity it named. The prompt is also equipped with examples of how the model can be wrong. These possible wrong answers were created from the previous runs and added on, as the database grew.

The prompt asks for four blocks, and the three that are not the findings themselves exist to make
refusals visible rather than silent: \texttt{considered} lists every model the paper touches with
a verdict on whether it was already released, \texttt{results\_covered} lists every result in the
paper and either points it at as a finding or says why it produced none, and
\texttt{concepts\_considered} explains, for a finding left untagged, which concept came closest and
why it did not apply. Concepts are chosen from a closed list of definitions carried in the prompt, so
an empty list is a valid answer.

Here is an example of the model output for a paper that produced one finding, abridged:
\begin{verbatim}
considered:
- model: GPT-4
  released: true
  why: OpenAI released model, publicly available via API, predates this paper
- model: Koala / imitation models (GPT-2 and Llama fine-tuned on ShareGPT-mix)
  released: false
  why: Models the authors trained by fine-tuning GPT-2 and Llama on ChatGPT
    outputs; they are the authors' own creations, not independently citable
    released checkpoints
[...]
findings:
- title: GPT-4, when used as a blind pairwise evaluator, exhibits the same
    style-over-factuality preference as human crowdworkers
  description: 'The paper uses GPT-4 to conduct blind pairwise comparisons
    between ChatGPT outputs and the authors'' imitation model outputs, following
    the same procedure as their Mechanical Turk evaluations. GPT-4''s ratings
    show the same trends as human raters [...]'
  evidence_type: correlational
  key_metric: ''
  caveat: The observation is brief and secondary to the paper's main focus on
    imitation models. No detailed analysis of GPT-4's evaluation mechanism is
    provided [...]
  models:
  - name: GPT-4
  concepts: []
  methods: []
  datasets: []
  related_work: []
results_covered:
- result: 'Crowdworker evaluation: imitation models rated ~70% as equal or
    better than ChatGPT'
  finding: none
  why: claim about the authors' imitation models, not a released model
- result: GPT-4 evaluation shows same trends as crowdworker evaluations
  finding: 1
[...]
entities:
- name: GPT-4
  kind: model
  citation: ''
  artifact: ''
concepts_considered:
- finding: GPT-4, when used as a blind pairwise evaluator, exhibits the same
    style-over-factuality preference as human crowdworkers
  closest: concept:shortcut
  why: The paper notes GPT-4's ratings correlate with human ratings that prefer
    style over factuality, but it does not demonstrate that GPT-4 specifically
    relies on style as a causal proxy for quality [...]
\end{verbatim}

\subsection{Step II: Entities}
After the answers are collected, split into multiple findings and citations verified, the names the extractor produced that no registry already holds are put to the model, one per prompt. The prompt asks whether the name earns a permanent entry in the shared registry, and it states
the stake plainly: an entry is how two papers come to talk about the same thing, so a wrong entry
either splits one thing into two records or merges two things into one.

Here we provide an example of the model output for two entities, one accepted and one rejected:
\begin{verbatim}
decision: adopt
why: "Named instruction-tuning dataset with its own repository, referenced as a
  data source by six unrelated papers."
title: "Stanford Alpaca"
anchor: "https://github.com/tatsu-lab/stanford_alpaca"

decision: refuse
why: "SGD is an optimiser, which is explicitly excluded from the registry."
\end{verbatim}

\subsection{Step III: Concepts and model attributes}
In the final step, we generate two prompts, both run over records that already exist in the database.

The first one is sent in order to match the correct concepts to findings that came from Step I without one. The model must quote the phrase of the definition that the finding satisfies. If it cannot quote one, the concept does not apply, and an empty answer is correct. 

Here is an example of model output to a finding about the model Sybil:
\begin{verbatim}
concepts:
- id: "concept:feature-interaction"
  because: "the definition covers cases where 'the output depends on a joint
    effect'; this finding tests exactly whether Sybil's output depends on
    pairwise (joint) effects between nodules versus single-nodule main effects,
    and reports that main effects suffice with only limited pairwise interactions"
\end{verbatim}

The second prompt is sent for every model entry in the registry and asks to fill in the following information about the entry model:
(1) \emph{Modality}, type of input the model takes or output it gives. (2) \emph{Task}, what it is for.
(3) \emph{Domain}, the field it was built for, recorded only when the model is specific to that
field. The first two accept one or more values. The model answers from the registry entry. In this case, an empty list is also a correct answer if the model cannot define an aspect.

Here is an example of model output:
\begin{verbatim}
modality:
- text
task:
- discriminative
domain: []
\end{verbatim}



\clearpage
\section{Extended analysis}
\label{app:extended-analysis}
\begin{figure}[h]
    \centering
    \begin{subfigure}[b]{0.48\textwidth}
        \centering
        \includegraphics[width=\textwidth]{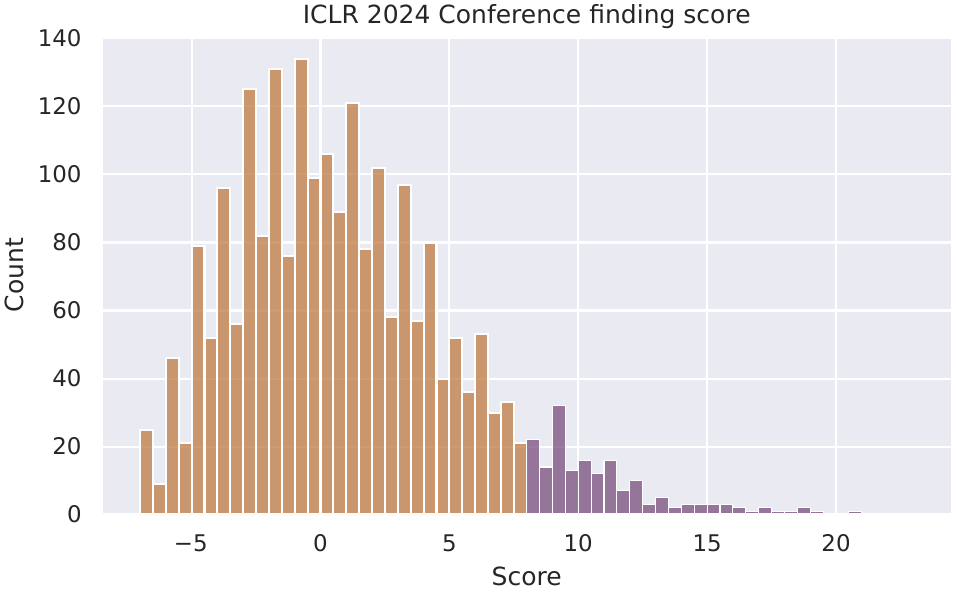}
    \end{subfigure}
    \hfill
    \begin{subfigure}[b]{0.48\textwidth}
        \centering
        \includegraphics[width=\textwidth]{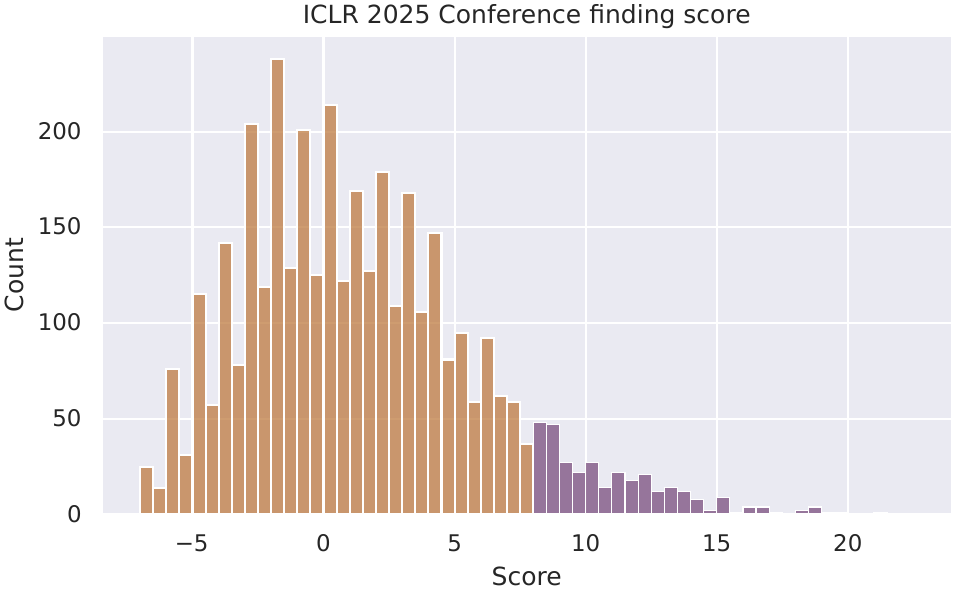}
    \end{subfigure}
    \caption{Distribution of conference finding score from ICLR 2024 (left) and ICLR 2025 (right). Papers scoring above the threshold are shown in purple, remaining papers in gold. Scoring method described in Appendix \ref{app:scoring-system}.}
    \label{fig:meta-analysis-1}
\end{figure}

\begin{figure}[h]
    \centering
    \begin{subfigure}[b]{0.48\textwidth}
        \centering
        \includegraphics[width=\textwidth]{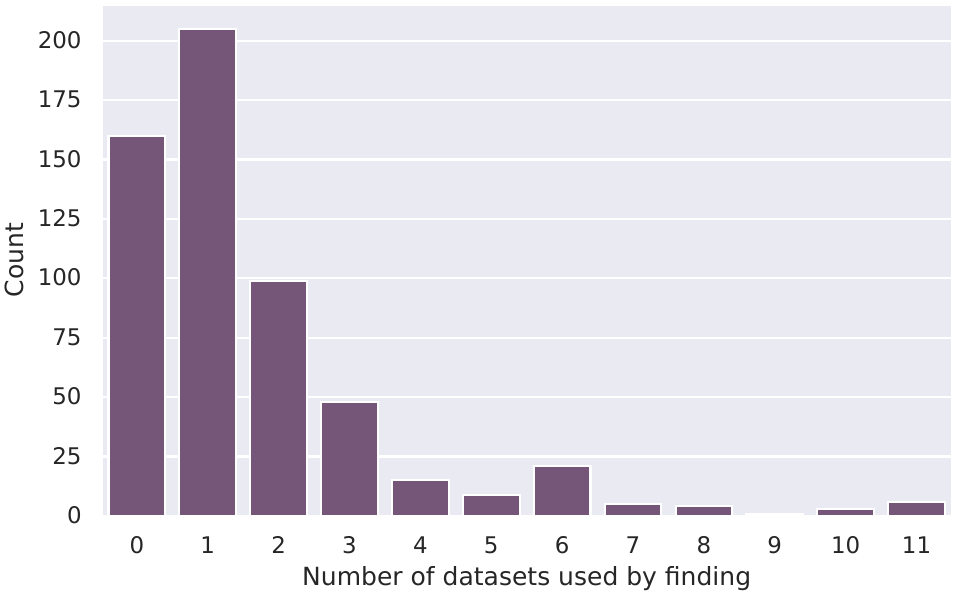}
    \end{subfigure}
    \hfill
    \begin{subfigure}[b]{0.48\textwidth}
        \centering
        \includegraphics[width=\textwidth]{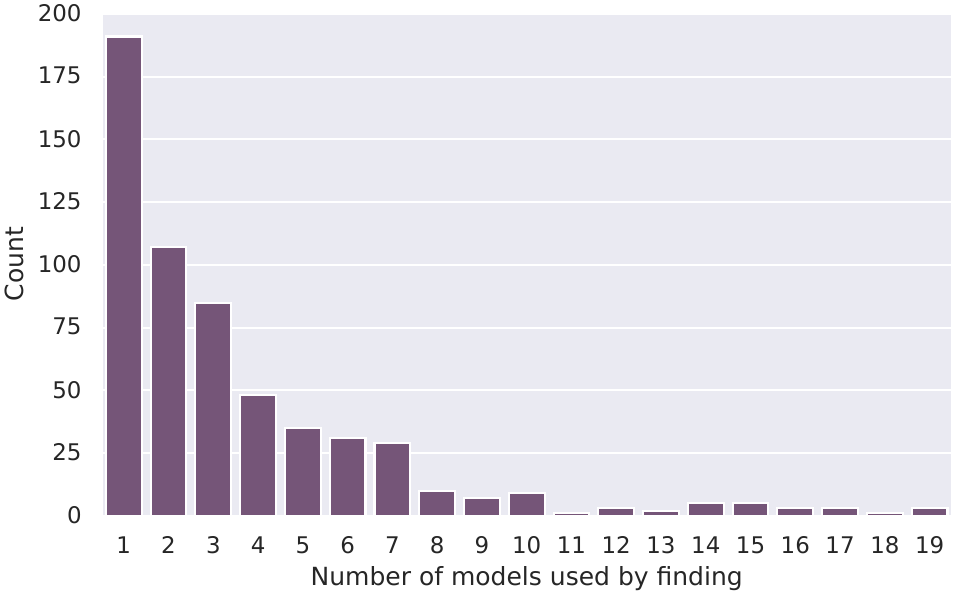}
    \end{subfigure}
    \caption{Distribution of number of datasets (left) and models (right) used per finding from ICLR 2025. Most findings rely on a single dataset and a single model, with long tails extending to 11 datasets and 19 models. Findings with zero datasets indicate that the authors either proposed a new dataset or the data is unavailable online.}
    \label{fig:meta-analysis-2}
\end{figure}

\begin{figure}[h]
    \centering
    \begin{subfigure}[b]{0.48\textwidth}
        \centering
        \includegraphics[width=\textwidth]{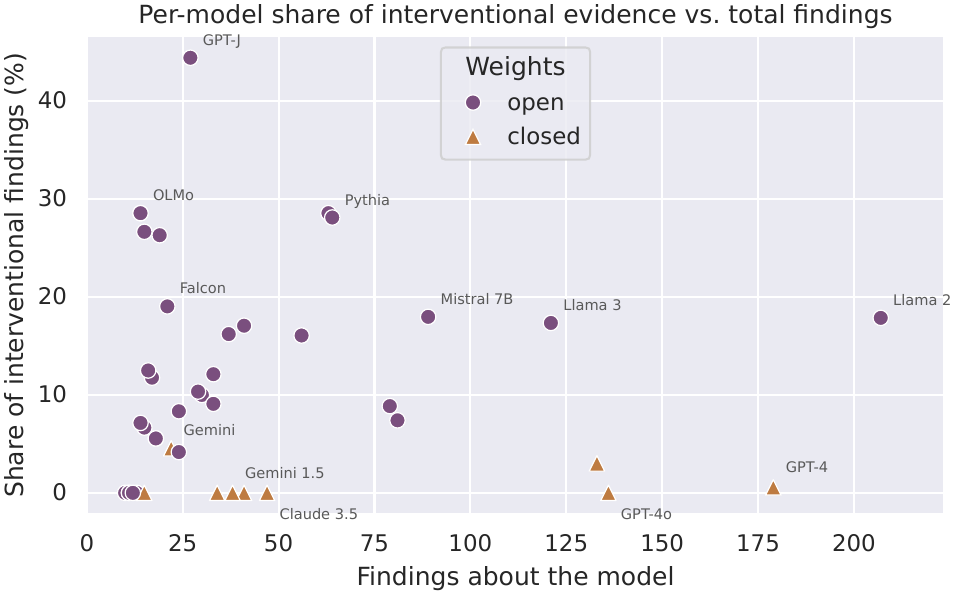}
    \end{subfigure}
    \hfill
    \begin{subfigure}[b]{0.48\textwidth}
        \centering
        \includegraphics[width=\textwidth]{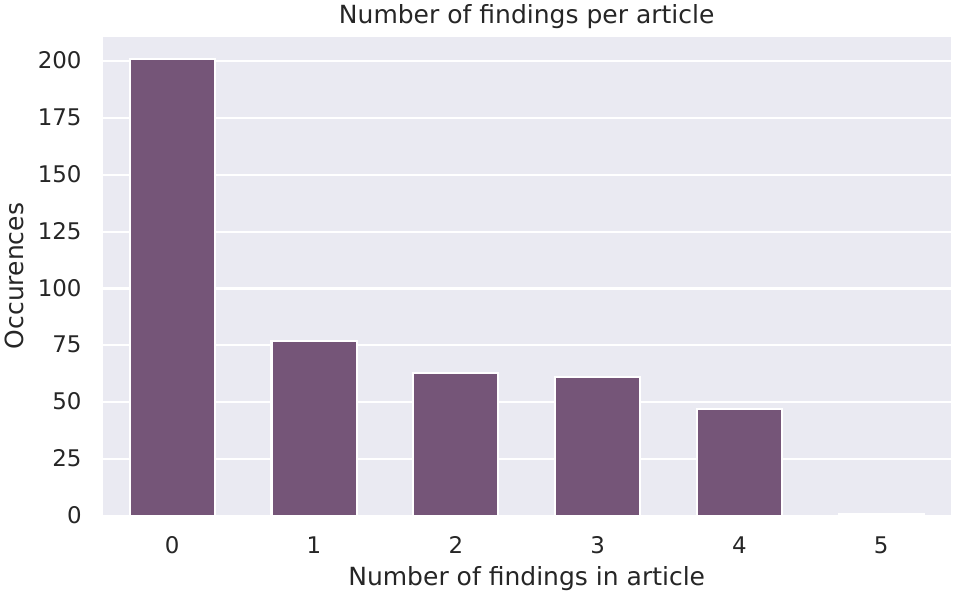}
    \end{subfigure}
    \caption{Interventional evidence share per model (left) for both considered conferences and extraction distribution per article (right) for ICLR 2025. Closed-weight models show near-zero interventional findings compared to the broader spread of open-weight models. Approximately $45\%$ of articles yielded no clear findings, with a maximum of 5 findings observed per article.}
    \label{fig:meta-analysis-3}
\end{figure}

In this section, we provide the additional results of the quantitative meta-analysis. Figure~\ref{fig:meta-analysis-1} presents the conference ranking score analysis from considered conferences, Figure~\ref{fig:meta-analysis-2} depicts the distribution of datasets and models contained in the extracted findings, and Figure~\ref{fig:meta-analysis-3} highlights the disparity in interventional evidence between open and closed-weight models, along with the distribution of findings per article.

\end{document}